\documentclass[a4paper,10pt]{article}
\pdfoutput=1
\usepackage[bookmarks=false]{hyperref}
\usepackage[numbers,sort&compress]{natbib}
\usepackage[margin=1in]{geometry}

\usepackage{cite}
\usepackage{color}
\usepackage{graphicx}
\usepackage{amsmath}
\usepackage{amssymb}
\usepackage{listings}
\usepackage[ruled, linesnumbered]{algorithm2e}
\usepackage[nonumberlist,acronym]{glossaries}
\usepackage{array}
\usepackage{placeins}
\usepackage{stfloats}
\usepackage{url}
\usepackage{multirow}
\usepackage{multicol}
\usepackage[table,xcdraw]{xcolor}
\usepackage[OT1]{fontenc}
\usepackage{epstopdf}
\usepackage{float}

\usepackage{mathtools}

\newcolumntype{P}[1]{>{\centering\arraybackslash}p{#1}}
\SetKwRepeat{Do}{do}{while}%

\usepackage{ctable}

\graphicspath{{./figures/}}

\newacronym{ROS}{ROS}{Robot Operating System}

\DeclareMathOperator{\diag}{diag}

\begin{document}

\newcommand{\coverTitle}{Indoor SLAM Using a Foot-mounted IMU and the local Magnetic Field}
\newcommand{\coverYear}{2022}

\newcommand{\coverAuthors}{Mostafa~Osman$^{\star}$, Frida~Viset$^{\star}$ and Manon~Kok$^{\star}$ \\ \vspace{3mm}
\small{$^\star$Delft Center for Systems and Control, Delft University of Technology, the Netherlands. \\e-mail: \{m.e.a.osman, f.m.viset, m.kok-1\}@tudelft.nl} 
}


\begin{titlepage}
\begin{center}
%

\vspace*{2.5cm}
%
{\Huge \bfseries \coverTitle  \\[0.4cm]}

%
{\Large \coverAuthors \\[1.5cm]}


\end{center}

\vspace{1cm}

\begin{abstract}
\noindent In this paper, a simultaneous localization and mapping (SLAM) algorithm for tracking the motion of a pedestrian with a foot-mounted inertial measurement unit (IMU) is proposed.
The algorithm uses two maps, namely, a motion map and a magnetic field map. 
The motion map captures typical motion patterns of pedestrians in buildings that are constrained by e.g. corridors and doors.
The magnetic map models local magnetic field anomalies in the environment using a Gaussian process (GP) model and uses them as position information.
These maps are used in a Rao-Blackwellized particle filter (RBPF) to correct the pedestrian position and orientation estimates from the pedestrian dead-reckoning (PDR). The PDR is computed using an extended Kalman filter with zero-velocity updates (ZUPT-EKF). 
The algorithm is validated using real experimental sequences and the results show the efficacy of the algorithm in localizing pedestrians in indoor environments. 
\end{abstract}


\vfill

\end{titlepage}

\title{Indoor SLAM Using a Foot-mounted IMU and the local Magnetic Field}
	
\author{Mostafa Osman$^{1}$, Frida Viset$^{1}$, Manon Kok$^{1}$ 
	\thanks{$^{1}$Delft Center for Systems and Control, Mechanical Maritime and Materials Engineering, Delft University of Technology, Delft, The Netherlands
		{\tt\small \{M.E.A.Osman, F.M.Viset, M.Kok-1\}@tudelft.nl}}%
}
\date{\empty}




\begin{figure}[t]
	\centering
	\includegraphics[trim={4cm 10.6cm 3.5cm 10.1cm}, clip, width=0.7\linewidth]{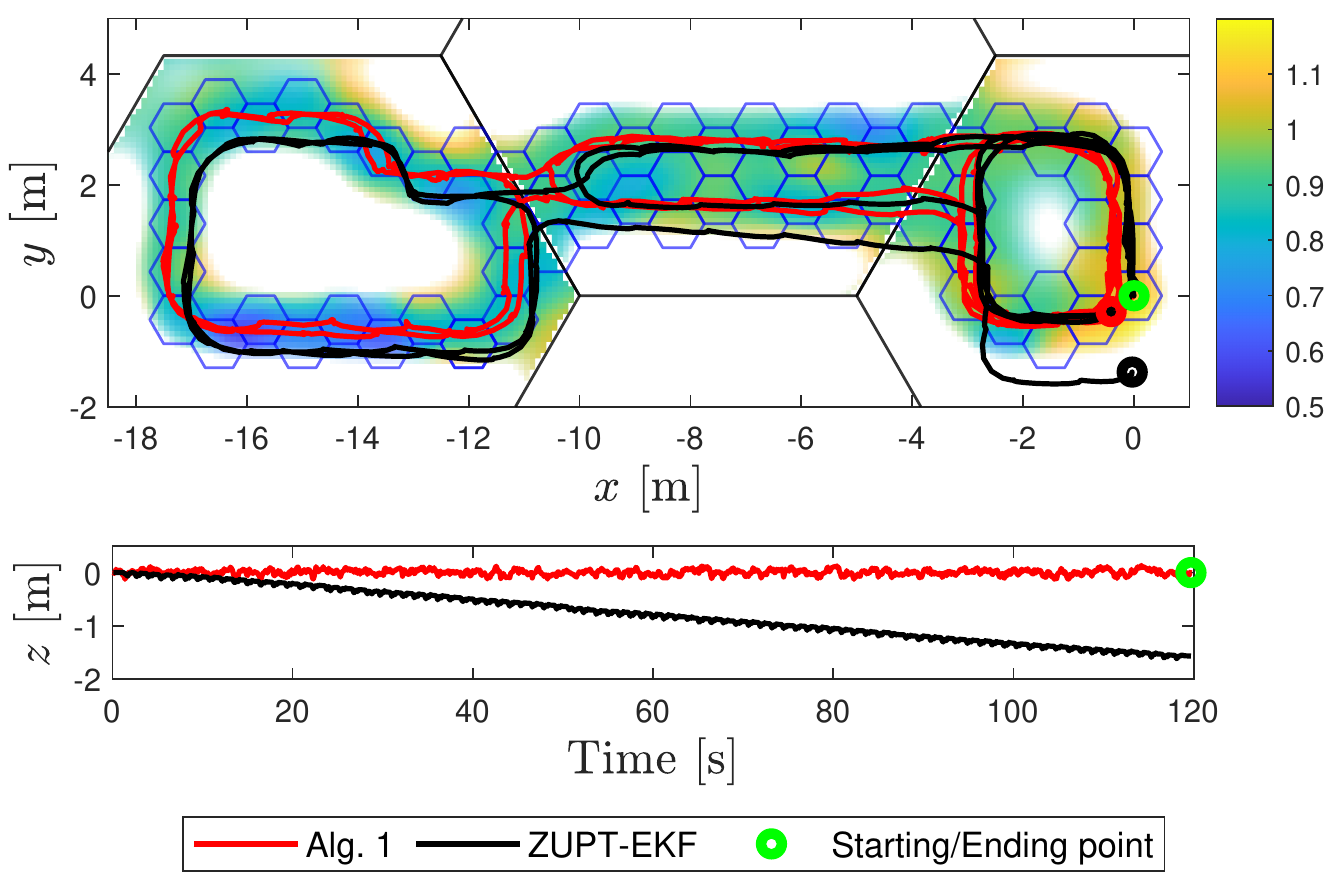}
	\caption{An example illustrating the performance of the proposed SLAM algorithm. Each of the black tiles (with radius 5 m and height 4 m) models a region of the magnetic field in the environment. The blue hexagons (with radius 0.5 m and height 0.25 m) models the typical directions of motion in the environment. The estimated path using the foot-mounted IMU and a ZUPT-EKF is shown in black and the estimated path by the proposed SLAM algorithm is shown in red. The actual starting (ending) point of the path marked using a green circular marker and the estimated end points of the ZUPT-EKF and the proposed SLAM algorithm are shown in black and red  circular markers respectively. The heat map shows the magnetic field norm.} \vspace{-0.3cm}
	\label{fig:PMSLAM}
\end{figure}

\section{Introduction}
\label{sec:Introduction}
Interest in indoor localization of pedestrians has increased in recent years due to the emergence of location-based services~\citep{he2017geomagnetism}. Indoor localization is a challenging task since Global Positioning System (GPS) signals get significantly weakened inside buildings. Additionally, relying on PDR approaches, where inertial measurements are integrated to estimate the pose, is infeasible due to the accumulation of drift even when using foot-mounted Inertial Measurement Units (IMU) with EKF aided with ZUPTs~\citep{nilsson2014foot, foxlin2005pedestrian, nilsson2012foot}. 

In order to solve the indoor localization problem without relying on GPS, several approaches were adopted in the literature using a variety of sensors such as cameras~\citep{garcia2016indoor, celik2009monocular}, Light Detection and Ranging (LiDAR)~\citep{peng2018improved, thrun2002probabilistic}, WiFi signals~\citep{bruno2011wislam, singh2018ensemble}, Ultra Wide Band (UWB)~\citep{stelios2008indoor}, and indoor magnetic field anomalies~\citep{kok2018scalable, kim2018rgb, jung2015magnetic, robertson2013simultaneous, viset2021magnetic}. In this work, we make use of the magnetic field anomalies as well as the typical motion of pedestrians inside buildings (shown in Fig.~\ref{fig:PMSLAM}). 

It is well-known that the earth is a dipole magnet that produces uniform magnetic flux. In indoor environments, the uniform magnetic field gets distorted due to the presence of metallic man-made objects. We use such magnetic distortions (anomalies) as a source of position information by building a map of the distorted magnetic field. 

In~\citep{kok2018scalable}, a SLAM algorithm was proposed which utilized the distorted magnetic field using an RBPF. Therein, reduced-rank Gaussian Process Regression (GPR)~\citep{solin2020hilbert} was used to map the magnetic field of the environment as described in~\citep{solin2018modeling}. 
Since the reduced rank GPR approximates the GP on a fixed domain~\citep{solin2020hilbert}, a partitioned map of 3D hexagonal tiles was used to construct a map of the magnetic field for large environments, thus making sure the computational cost of the GPR does not scale with the size of the environment. Furthermore, an RBPF exploits a conditionally linear substructure making the particle filter also tractable for large state dimensions.

A different approach to solving the pedestrian indoor localization problem using only a foot-mounted IMU was proposed in~\citep{angermann2012footslam}, namely, the FootSLAM algorithm. The FootSLAM algorithm aims at modeling the motion of a pedestrian inside a building. Such motion is typically governed by the structure of the building such as the presence of corridors, walls, doors, and so on. This was accomplished through defining a 2D map of hexagonal tiles, each edge of the hexagonal tile having a probability of being crossed. An RBPF was used to estimate the pose of the pedestrian while updating the motion map. The importance weights were calculated based on the 2D hexagonal tiles map~\citep{angermann2009inertial}. 

In this paper, we propose a SLAM algorithm that uses both a magnetic field map and a motion map of the pedestrian motion patterns. These two sources of information (magnetic field and pedestrian motion) are complementary and hence result in more accurate localization and mitigate some of the drawbacks of each approach such as the absence of local magnetic field anomalies or moving in wide areas with no motion patterns. This approach was first introduced in~\citep{robertson2013simultaneous} where the magnetic field map representation was a multi-layer hexagonal grid map. Each layer had an increased resolution of the magnetic field of the environment. Here, we adopt a more continuous magnetic field model by using the reduced rank GPR with 3D hexagonal tiles as in~\citep{kok2018scalable}. Furthermore, we extend the motion map to 3D to accurately estimate the 3D motion of the pedestrian's foot and the magnetic field in 3D space. The algorithm is validated using several real experimental scenarios. Quantitative evaluation of the algorithm is accomplished using the ground-truth provided by an optical motion capture system. 

The remainder of this paper is organized as follows: Section~\ref{sec:probForm} contains the motion model as well as the magnetic field model and the definition of the motion maps followed by the algorithm details in Section~\ref{sec:algorithm}. The experimental results and evaluation of the proposed approach are presented and discussed in Section~\ref{sec:experimental_work}. Finally, Section~\ref{sec:conclusion} contains some concluding remarks as well as some suggestions for the future work. 

\section{Modeling}
\label{sec:probForm}

The SLAM problem we tackle in this paper is the well-known problem of approximating the distribution~\citep{osman2019intelligent}
\begin{equation}
\label{equ:main_posterior}
p(\mathbf{x}_{1:k}, M \mid \mathbf{u}_{1:k}, \mathbf{z}_{1:k}),    
\end{equation}
where $\mathbf{x} := [\mathbf{p}^\top, \mathbf{q}^\top]^{\top}$ is the pose vector consisting of the three-dimensional position $\mathbf{p} \in \mathbb{R}^{3}$ and the orientation $\mathbf{q} \in \mathbb{M} \subset \mathbb{R}^{4}$ represented in unit quaternions, $M := \{M_{\text{m}}, M_{\text{u}}\}$ is the environment map consisting of the magnetic field map $M_\text{m}$ and the pedestrian motion map $M_\text{f}$. Furthermore, $\mathbf{u} := [\Delta \mathbf{p}^{\top}, \Delta \mathbf{q}^{\top}]^{\top} \in \mathbb{R}^{7}$ is the motion increment estimated using a foot-mounted IMU, $\mathbf{z} := [(\mathbf{z}^{\text{m}})^\top, (\mathbf{z}^{\text{u}})^\top]^{\top}$ is the measurement vector consisting of the face transition measurements in the hexagons of the motion map $\mathbf{z}^\text{u} \in \mathbb{R}^{6}$ and the magnetic field measurement $\mathbf{z}^{\text{m}} \in \mathbb{R}^{3}$ captured using a magnetometer, and $k \in \mathbb{N}$ is the time-step. Notice that here we assume that the environment is static and consequently $M$ is constant.

In this paper, an omni-directional motion model subjected to the zero-mean Gaussian white noises $\varepsilon_{k}^{\text{p}} \in \mathbb{R}^{3}$ and $\varepsilon^{\text{q}}_{k} \in \mathbb{R}^{3}$  is used to model the pedestrian foot motion as
\begin{equation}
	\label{equ:motion_model}
	\mathbf{x}_{k} = \begin{bmatrix}
		\mathbf{p}_{k-1} + \Delta \mathbf{p}_{k} + \varepsilon_{k}^{\text{p}} \\
		\Delta \mathbf{q}_{k} \odot \mathbf{q}_{k-1} \odot \exp_{\text{q}}(\varepsilon^{\text{q}}_{k})
	\end{bmatrix}, \ \ \begin{bmatrix}
		\varepsilon_{k}^{\text{p}} \\ \varepsilon^{\text{q}}_{k}
	\end{bmatrix} \sim \mathcal{N}(\mathbf{0}, \mathcal{Q}).
\end{equation}
where $\exp_{\mathbf{q}} : \mathbb{R}^{3} \rightarrow \mathbb{M}$ is the exponential map from the tangent space to the quaternion space, and $\odot$ is the quaternion product (for more information, see~\citep{kok2017inertial}). 
Moreover, the orientation $\mathbf{q}_{k}$ is the orientation of the sensor frame with respect to the world frame. The origin of the sensor frame lies in the centre of the
magnetometer triad and its axes are aligned with the axes of the magnetometer.

The magnetic field measurements are modeled as the gradient of a scalar potential field as proposed in~\citep{solin2018modeling} as 
\begin{equation}
	\mathbf{z}^{\text{m}}_{k} = R(\mathbf{q}_{k})^{\top} \nabla \varphi(\mathbf{p}_{k}) + \varepsilon^{\text{m}}_{k}.
\end{equation}
where $\varphi : \mathbb{R}^{3} \rightarrow \mathbb{R}$ is the scalar potential field, $R: \mathbb{M} \rightarrow SO(3)$ maps from the quaternion representation to the corresponding rotation matrix, and  $\varepsilon_{k}^{\text{m}} \sim \mathcal{N}(0, \mathcal{R}^{\text{m}})$ is a Gaussian zero-mean white measurement noise. 

The motion of the pedestrian is captured by using 3D hexagonal partitioning of the space of motion. The transition measurement $\mathbf{z}_{k}^{\text{u}}$ is a detector of the pedestrian crossing from one hexagon to another through one of the faces of the hexagonal tiles, i.e. $\mathbf{z}_{k}^{\text{u}}$ determine if and which face of the 3D hexagonal tiles in the motion map was crossed.
  
Now, going back to the posterior in~\eqref{equ:main_posterior}, it can be factored as 
\begin{equation}
\label{equ:factored_posterior}
    p(\mathbf{x}_{1:k}, M \mid \mathbf{z}_{1:k}, \mathbf{u}_{1:k}) = 
    \underbrace{p(M \mid \mathbf{x}_{1:k}, \mathbf{z}_{1:k}, \mathbf{u}_{1:k})}_{\text{Map posterior with know poses}} \overbrace{p(\mathbf{x}_{1:k} \mid \mathbf{z}_{1:k}, \mathbf{u}_{1:k})}^{\text{Pose posterior}}.
\end{equation}

By applying Bayes' rule, and given the motion model and the measurement models used, the pose posterior can be written as 
\begin{align*}
    p(\mathbf{x}_{1:k} \mid \mathbf{z}_{1:k}, \mathbf{u}_{1:k}) & = \eta \  p(\mathbf{z}_{k} |\mathbf{x}_{1:k}, \mathbf{u}_{1:k}, \mathbf{z}_{1:k-1}) \\ 
    & \ \ \ \ \ p(\mathbf{x}_{1:k} | \mathbf{u}_{1: k}, \mathbf{z}_{1: k-1}) \\
    &= \eta \ p(\mathbf{z}_{k} \mid \mathbf{x}_{k}, \mathbf{u}_{k}) p(\mathbf{x}_{k} \mid \mathbf{x}_{k-1}, \mathbf{u}_{k}) \\ 
    & \ \ \ \ \ p(\mathbf{x}_{1:k-1} \mid \mathbf{u}_{1: k-1}, \mathbf{z}_{1: k-1}) \\
    & = \eta \ p(\mathbf{z}^{\text{m}}_{k} \mid \mathbf{x}_{k})p(\mathbf{z}^{\text{u}}_{k} \mid \mathbf{x}_{k}, \mathbf{u}_{k}) p(\mathbf{x}_{k} \mid \mathbf{x}_{k-1}, \mathbf{u}_{k}) \\ 
    & \ \ \ \ \  p(\mathbf{x}_{1:k-1} \mid \mathbf{u}_{1: k-1}, \mathbf{z}_{1: k-1}),
\end{align*}
where $\eta$ is the measurement marginal distribution and is treated as a normalizing constant. 

Since the map $M_{\text{m}}$ of the magnetic field and the map $M_{u}$ of the pedestrian motion are independent, the map posterior can be written as 
\begin{equation}
p(M \mid \mathbf{x}_{k}, \mathbf{z}_{1:k}, \mathbf{u}_{1:k}) = p(M_\text{m} \mid \mathbf{x}_{k}, \mathbf{z}^{\text{m}}_{k})p(M_\text{u} \mid \mathbf{x}_{k}, \mathbf{z}^{\text{u}}_{k}). 
\end{equation}

In the following two subsections, the modeling of the magnetic field mapping as well as the motion map are explained. 

\subsection{Magnetic field mapping}
In this paper, we use GPR to represent the magnetic field map $M_\text{m}$. The indoor magnetic field is modeled as the gradient of a curl-free scalar potential field $\varphi$ and the GP prior is defined as 
\begin{equation}
	\label{equ:fullGPR}
	\varphi(\mathbf{p}) \sim \mathcal{G}\mathcal{P}(0, \kappa_{\text{lin.}}(\mathbf{p}, \mathbf{p}') + \kappa_{\text{SE}}(\mathbf{p}, \mathbf{p'})).
\end{equation}

The earth's contribution to the local magnetic field is captured in the GP prior using the linear covariance function (kernel)
\begin{equation}
	\label{equ:linear_kernel}
	\kappa_{\text{lin.}} (\mathbf{p}, \mathbf{p}') := \sigma^{2}_{\text{lin.}} \mathbf{p}^{\top} \mathbf{p}',
\end{equation}
while the local magnetic field anomalies are assumed to follow a squared exponential kernel
\begin{equation}
	\label{equ:se_kernel}
	\kappa_{\text{SE}}(\mathbf{p}, \mathbf{p}') := \sigma_{\text{SE}}^{2} \exp\left(-\frac{||\mathbf{p} - \mathbf{p}'||^2_2}{2\ell^{2}}\right),
\end{equation}
where $\sigma_{\text{lin}} \in \mathbb{R}$ and $\sigma_{\text{SE}} \in \mathbb{R}$ are the linear kernel and the squared exponential kernel magnitude scales, respectively, and $\ell \in \mathbb{R}$ is the length scale.

Full GP regression to estimate the magnetic field map is intractable in practice since it relies on inverting the covariance matrix which is an operation with $\mathcal{O}(N^3)$ complexity where $N$ is the number of measurements. Consequently, similar to~\citep{kok2018scalable}, we use the reduced rank GP regression proposed in~\citep{solin2020hilbert}. 
The GP regression problem is projected on the eigenbasis of the (negative) Laplace operator in a confined domain $\Omega \subset \mathbb{R}^{3}$. In this domain, the eigendecomposition of the Laplace operator subject to the Dirichlet boundary conditions can be solved as
\begin{align}
	\label{equ:laplacian_equation}
	\left\{
	\begin{array}{cl}
		-\nabla^2 \phi_{j}(\mathbf{p}) &= \lambda^{2}_{j}\phi_{j}(\mathbf{p}), \ \ \ \mathbf{p} \in \Omega, \\
		\phi_{j}(\mathbf{p}) & = 0, \ \ \ \ \ \ \ \ \ \  \  \mathbf{p} \in \partial\Omega. 
	\end{array}
	\right.
\end{align}

This eigendecomposition problem enables the approximation of the covariance matrix in~\eqref{equ:fullGPR} as
\begin{equation}
	\kappa(\mathbf{p}, \mathbf{p}') := \kappa_{\text{lin.}}(\mathbf{p}, \mathbf{p}') + \kappa_{\text{SE}}(\mathbf{p}, \mathbf{p}') 
	 \approx \kappa_{\text{lin.}}(\mathbf{p}, \mathbf{p}') + \sum_{j=1}^{m} S_{\text{SE}}( \lambda_{j})\phi_{j}(\mathbf{p})\phi_{j}(\mathbf{p}'), 
\end{equation}
where $\{\phi_{j}, \lambda_{j}\}_{j = 1}^{m}$ are the eigenfunctions and their corresponding eigenvalues, and $S_{\text{SE}}(\cdot)$ is the spectral density of the squared exponential kernel.

As in~\citep{solin2018modeling}, we use the GP regression approximation described to model the magnetic field. This approximation is known to be exact for infinite sizes of the domain and infinitely many basis functions. To model large spaces with smaller number of basis functions, making the representation more scalable, as in~\citep{kok2018scalable}, we split the map into three-dimensional hexagonal blocks, each modeled as a separate GP. Each of such hexagonal blocks is defined as
\begin{equation}
\Omega_{\text{m}}^{(d)} := \{\mathbf{p} \mid (x, y) \in \text{hexagon}(r^{\text{m}}, (x_c, y_c)^{\text{m}, (d)}), 
z \in [z^{\text{m}, (d)}_{c} - L^{\text{m}}_z, z^{\text{m}, (d)}_{c} + L^{\text{m}}_z]\},
\end{equation}
where $r^{\text{m}}$ and $(x_c, y_c)^{\text{m}, (d)}$ are the radius and the center of the hexagon respectively, $L^{\text{m}}_z$ is half the height of the hexagon, the position vector is decomposed as $\mathbf{p} := [x, y, z]^{\top} \in \Omega_{\text{m}}^{(d)}$, and the operating space is defined as $\Omega_{\text{m}} := \bigcup_{d = 1}^{N_\text{m}} \Omega^{(d)}_{\text{m}}$. For the details of computing the eigenfunctions and eigenvalues, see~\citep{kok2018scalable}.

\subsection{Pedestrian motion mapping}
As for the pedestrian motion map, the map $M_\text{u}$ 
captures typical motion patterns of pedestrians in buildings  which are constrained by e.g. corridors and doors.
This is accomplished through discretizing the space into 3D hexagonal tiles, each of the hexagonal tile faces has a given transition probability. 
Each hexagon $\Omega^{(d)}_{\text{u}}$ is defined similar to the magnetic map as 
\begin{equation}
	\Omega_{\text{u}}^{(d)} := \{\mathbf{p} \mid (x, y) \in \text{hexagon}(r^{\text{u}}, (x_c, y_c)^{\text{u}, (d)}), z \in [z_{c}^{\text{u}, (d)} - L^{\text{u}}_z, z_{c}^{\text{u}, (d)} + L^{\text{u}}_z]\},
\end{equation}
and the full indoor environment is defined as $\Omega_\text{u} := \bigcup_{i = 1}^{N_\text{u}} \Omega_{\text{u}}^{(i)}$. 

For each hexagon, the pedestrian motion map $M_{\text{u}}^{(d)} \in [0, 1]^{8}$ encodes the transition probabilities of all the faces
\begin{equation}
	\label{equ:transition_prob}
	M_{\text{u}, j}^{(d)} := P(\mathbf{p}_{k} \in \Omega_{\text{u}}^{(d)} \mid \mathbf{p}_{k-1} \in \Omega_{\text{u}}^{(\tilde{d})}, f_{j} \in \Omega_{\text{}u}^{(\tilde{d})}), 
\end{equation}
where $f_{j} \in \Omega^{(d)}_{\text{u}}$ is the $j$-th face of the $d$-th hexagon and $\Omega_{\text{u}}^{(\tilde{d})}$ is a neighboring hexagon. 

\section{Pedestrian Magnetic SLAM}
\label{sec:algorithm}
In this section, our solution to the estimation problem described in~\eqref{equ:factored_posterior} is stated. First, we state the Rao-blackwellized PF formulation, followed by the estimation approaches of the magnetic map and the pedestrian motion map. 

\subsection{Rao-Blackwellized particle filter}
We use a Rao-blackwellized PF to estimate the posterior in~\eqref{equ:main_posterior}. The pose posterior shown in~\eqref{equ:factored_posterior} is approximated by a set of particles $\{\mathbf{x}^{[i]}\}_{i = 1}^{N_{\text{p}}}$ each with an impulse distribution
\begin{equation}
	\{\mathbf{x}_{k}^{[i]}\}_{i=1}^{N_\text{p}} \sim p(\mathbf{x}_{1:k} \mid \mathbf{z}_{1:k}, \mathbf{u}_{1:k}).
\end{equation}

Through Rao-Blackwellization~\citep{murphy2001rao}, a sample of the map posterior can be computed for each of the sampled particles. This is accomplished by assigning to each of the sampled particles an estimate of the map $M$ conditioned on the particle $p(M^{[i]} \mid \mathbf{x}^{[i]}_{k}, \mathbf{z}_{k}) = p(M_{\text{m}}^{[i]} \mid \mathbf{x}^{[i]}_{k}, \mathbf{z}^{\text{m}}_{k})p(M_{\text{u}}^{[i]} \mid \mathbf{x}^{[i]}_{k}, \mathbf{z}_{k}^{\text{u}})$. Thus approximating the full joint posterior stated in~\eqref{equ:factored_posterior}.

The particle set is drawn from a proposal distribution $\pi(\mathbf{x}_{k} \mid \mathbf{u}_{1:k}, \mathbf{z}_{1:k})$. In this paper, we use a bootstrap particle filter, hence, the proposal distribution used is 
\begin{equation}
\pi(\mathbf{x}_{1:k} | \mathbf{u}_{1:k}, \mathbf{z}_{1:k}) := p(\mathbf{x}_{k} | \mathbf{x}_{k-1}, \mathbf{u}_{k}).
\end{equation}
which can be sampled using the model stated in~\eqref{equ:motion_model}. To be able to actually approximate the target distribution using the sampled particles, an importance weight is assigned to each particle which is computed using~\eqref{equ:factored_posterior} as
\begin{equation}
	\label{equ:weight_calc}
w^{[i]}_{k} := \frac{p(\mathbf{x}_{1:k} \mid \mathbf{z}_{1:k}, \mathbf{u}_{1:k})}{\pi(\mathbf{x}_{1:k} \mid \mathbf{u}_{1:k}, \mathbf{z}_{1:k})} 
\ = \eta \ \underbrace{p(\mathbf{z}^{\text{m}}_{k} \mid \mathbf{x}_{k})}_{w_{k}^{\text{m},[i]}} \underbrace{p(\mathbf{z}^{\text{u}}_{k} \mid \mathbf{x}_{k}, \mathbf{u}_{k})}_{w_{k}^{\text{u}, [i]}} w_{k-1}^{[i]}.
\end{equation}

The particles are re-sampled with replacement with a probability proportional to the importance weights~\citep{hol2006resampling}. 
The pose posterior can be more formally represented using the resampled particle-weight set 
\begin{equation}
    \{<\mathbf{x}_{k}^{[i]}, w^{[i]}_{k}>\}_{i=1}^{N_\text{p}} \sim p(\mathbf{x}_{1:k} \mid \mathbf{z}_{1:k}, \mathbf{u}_{1:k}).
\end{equation}

It was previously mentioned that each particle has an estimate of the maps $M_{\text{m}}, M_{\text{u}}$. 
Using the reduced rank GP regression approach described above, it is possible to write the prior for each magnetic hexagonal tile in terms of a mean $\mathbf{m}^{(d)}_{0} = \mathbf{0}_{m+3}$ and a covariance matrix
\begin{equation}
    P_{0}^{(d)} := \mathbf{diag}(\sigma_{\text{lin.}}^{2}, \sigma_{\text{lin.}}^{2}, \sigma_{\text{lin.}}^{2}, S_{\text{SE}}(\lambda_{1}), \dots, S_{\text{SE}}(\lambda_{m})).
\end{equation}

The posterior $p(M_{\text{m}}^{(d)}|\mathbf{x}_{k}^{[i]}, \mathbf{z}^{\text{m}}_{k}) \sim \mathcal{N}(\mathbf{m}^{[i], (d)}_{k}, P_{k}^{[i], (d)})$ can consequently be recursively computed for a particle with position $\mathbf{p}^{[i]}_{k} \in \Omega_{\text{m}}^{(d)}$ using the Kalman filter measurement update step as in~\citep{solin2018modeling} 
\begin{align}
\label{equ:mag_update1}
\begin{split}
	\mathcal{S}_{k} &= C^{[i]}_{k} P_{k-1}^{[i], (d)} (C^{[i]}_{k})^{\top} + \mathcal{R}^{m}, \\
	\mathcal{K}_{k} &= P_{k-1}^{[i], (d)} (C^{[i]}_{k})^{\top} \mathcal{S}_{k}^{-1}, \\
	\mathbf{m}^{[i], (d)}_{k} &= \mathbf{m}^{[i], (d)}_{k-1} + \mathcal{K}_{k}(\mathbf{z}_{k}^{\text{m}} - C^{[i]}_{k} \mathbf{m}^{[i], (d)}_{k-1}), \\
	\mathbf{P}_{k}^{[i], (d)} &= P_{k-1}^{[i], (d)} - \mathcal{K} \mathcal{S}_{k} \mathcal{K}_{k}^{\top},
\end{split}
\end{align}
where $C^{[i]}_{k} := R(\mathbf{q}^{[i]}_{k})^{\top} \nabla \boldsymbol\Phi_{k}^{[i]}$ is the measurement matrix, 
$\mathcal{S}_{k} \in \mathbb{R}^{3 \times 3}$ is the innovation covariance matrix, $\mathcal{K}_{k} \in \mathbb{R}^{(m+3) \times 3}$ is the Kalman gain, and $\nabla \boldsymbol\Phi_{k}^{[i]} \in \mathbb{R}^{3 \times (m+3)}$ is the gradient of the basisfunctions evaluated at the position of the $i$-th particle
\begin{equation}
	\label{equ:magmapend}
	\nabla \boldsymbol\Phi_{k}^{[i]} := ((\nabla \mathbf{p}^{[i]}_{k})^{\top}, \nabla \phi_{1}(\mathbf{p}^{[i]}_{k}), \dots, \nabla \phi_{m}(\mathbf{p}^{[i]}_{k})).
\end{equation}


To avoid the boundary effects of the Dirichlet boundary conditions shown in~\eqref{equ:laplacian_equation}, the size of the magnetic hexagonal tiles is slightly extended outside the configured sizes of the tiles. This helps build continuous maps of the magnetic field using the hexagon tiles since the partitioned domains of the GPs overlap. If a particle is located in these overlapping regions, then all hexagonal tiles that contain the location are updated using magnetic measurement. 

As for the motion map, the map is updated by incrementing the number of transitions for a given face when it is crossed and the transition probabilities defined in~\eqref{equ:transition_prob} is computed as 
\begin{equation}
	\label{equ:footmaph}
	M_{\text{u}, j}^{(d)} = \begin{cases}
		\frac{N_{j}^{(d)}}{\sum_{j = 1}^{8} N^{(d)}_{j}} & \text{if $j \in \{1, \dots 6\}$} \\
		\ \ \ \ p_{v} & \text{if $j \in \{7, 8\}$}
	\end{cases},
\end{equation}
where $N_{j}^{(d)}$ is the total number of transitions occurred through the face $j$. For the upper and lower faces ($j=\{7, 8\}$) of the hexagon, we assign a small transition probability $p_{v}$ to account for the improbability of vertical motion for pedestrians. 

In order to make sure the particle cloud does not collapse very quickly during operation and to avoid particle degeneracy, the resampling of the particles is only done when the effective size of the particle cloud goes under a certain threshold
\begin{equation}
	N_{\text{eff}} := \frac{1}{\sum_{i=1}^{N_p} \left(w^{[i]}_{k}\right)^{2}} < \frac{3}{4}N_{\text{p}},
\end{equation}
where $N_{\text{eff}}$ is the effective sample size.

Our SLAM algorithm builds the maps of the environment with respect to the initial pose of the pedestrian's foot. Therefore, it is possible to move or rotate the estimated maps while simultaneously moving or rotating the initial pose, i.e., the absolute position and orientation are not observable. Consequently, we align the initial position and orientation of the pedestrian's foot to the world frame. The proposed SLAM algorithm is summarized in Algorithm~\ref{alg:LSMHE} where $\mathbf{0}_{n}, $$\mathbf{1}_{n}$ denotes all zeros and all ones n-vectors respectively and $\mathbf{q}_{0} = [1, \mathbf{0}_{3}^{\top}]$ denotes the quaternion of zero rotation.

\begin{algorithm}[t]
	\small
	\SetKwData{Left}{left}\SetKwData{This}{this}\SetKwData{Up}{up}
	\SetKwFunction{Union}{Union}\SetKwFunction{FindCompress}{FindCompress}
	\SetKwInOut{Input}{Input}\SetKwInOut{Output}{Output}
	
	\Input{$\mathbf{z}_{k}^{\text{m}}, \mathbf{u}_{k}$, for $k = 1 \dots N_\text{T}$, $\mathcal{Q}$, $\mathcal{R}$, $m$, $\sigma_{\text{SE}}$, $\sigma_{\text{lin}}$, $\ell$, $r^{\text{m}}$, $r^{\text{u}}$, $L_{z}^{\text{m}}$, $L_{z}^{\text{u}}$, and $N_{\text{p}}$.} 
	\Output{For each timestep $k$, the estimated $\mathbf{x}_{k}$, $M_\text{m}$, $M_\text{u}$ of the highest weight particle.}
	\BlankLine
	\For{$ i \in \{1 \dots N_{p}\}$}{
		\textit{Initialize} the particles as $\mathbf{x}_{0}^{[i]} = [\mathbf{0}_{3}^{\top}, \mathbf{q}^{\top}_{0}]^{\top}$ and the importance weights as $w_{0}^{[i]} = \frac{1}{N_{p}}$\\ 
		\textit{Initialize} the maps 
		\begin{equation*}
			M^{[i]}_{m} \leftarrow \mathcal{N}(\mathbf{0}_{m+3}, \mathbf{P}_{0}^{[i], (1)})
		\end{equation*}
		\begin{equation*}
			M^{[i]}_{u} \leftarrow (M_{u}^{[i], (1)} = \mathbf{1}_{8})
		\end{equation*}
	}
	\For{$k \in \{1 \dots N_T\}$}
	{ 
		\For{$ i \in \{1 \dots N_{p}\}$}{
			\textit{Propagate} each particle using the motion model according to~\eqref{equ:motion_model}. \\
			\textit{Check} if the new particle pose is in a new magnetic (motion) hexagonal tile. If yes, initialize the new tile and add it to the magnetic (motion) map. \\
			\textit{Determine} $\mathbf{z}^{\text{u}}_{k}$ for the particle based on the change in the particle position. \\
			\textit{Compute} the importance weight $w_{k}^{\text{m}, [i]}$ using the GP.\\
			\textit{Assign} the importance weight $w_{k}^{\text{u}, [i]}$ using corresponding probabilities computed according to~\eqref{equ:footmaph}.\\
			\textit{Compute} the importance weights for each particle according to~\eqref{equ:weight_calc}. \\
			Update the maps $M_{m}$ using~\eqref{equ:mag_update1} and $M_{u}$ by updating $N_{j}^{(d)}$. \\
		}
		
		\textit{Normalize} the importance weights. \\
		\If{$N_{eff} < \frac{3}{4} N_{\text{p}}$}{
		\textit{Resample} the particles (with replacement) with probabilities proportional to the weights. \\ 	
		}	
	}
	\caption{Pedestrian Magnetic SLAM} 
	\label{alg:LSMHE}
\end{algorithm}


\section{Experimental Results}
\label{sec:experimental_work} 
To evaluate the performance of Alg.~\ref{alg:LSMHE}, two data sets were executed with the ground-truth recorded using an Optitrack motion capture system in the limited space of a motion capture lab as well as a longer data set collected without ground-truth data. 
In order to accurately evaluate the performance of the proposed algorithm and to capture the stochasticity of the particle filter, Algorithm 1 was run 10 times for each data set. For both data sets we compute the average RMSE in both position and orientation over these 10 runs. 
As for the long data set, since no ground-truth data is available, we designed the path such that the end point is the same as the start point to validate the performance of the algorithm by the deviation between the end point estimated by the algorithm and the known starting point. 

\begin{figure}[t]
	\centering
	\begin{tabular}{cc}
		\includegraphics[width=0.3\linewidth]{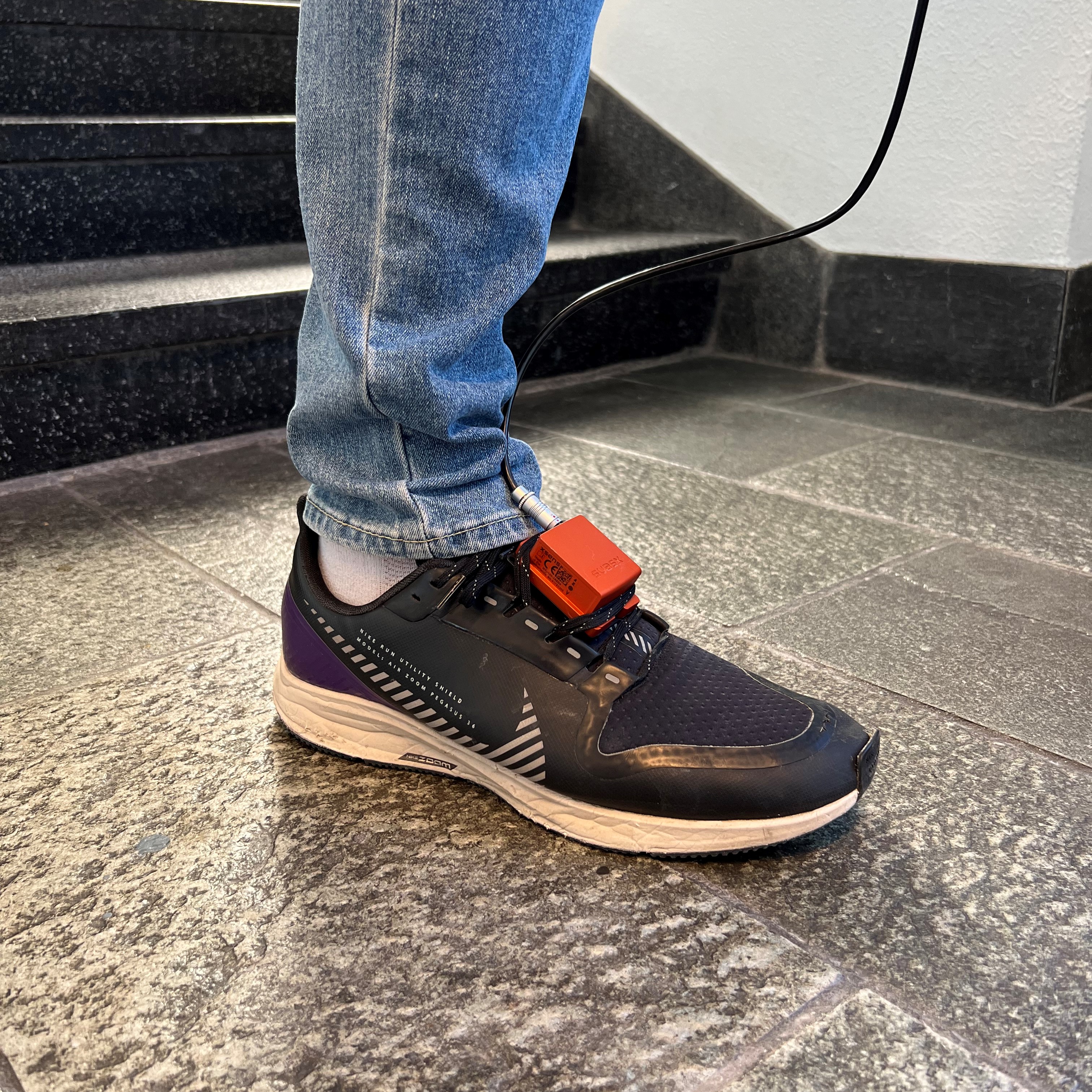} &
		\includegraphics[width=0.3\linewidth]{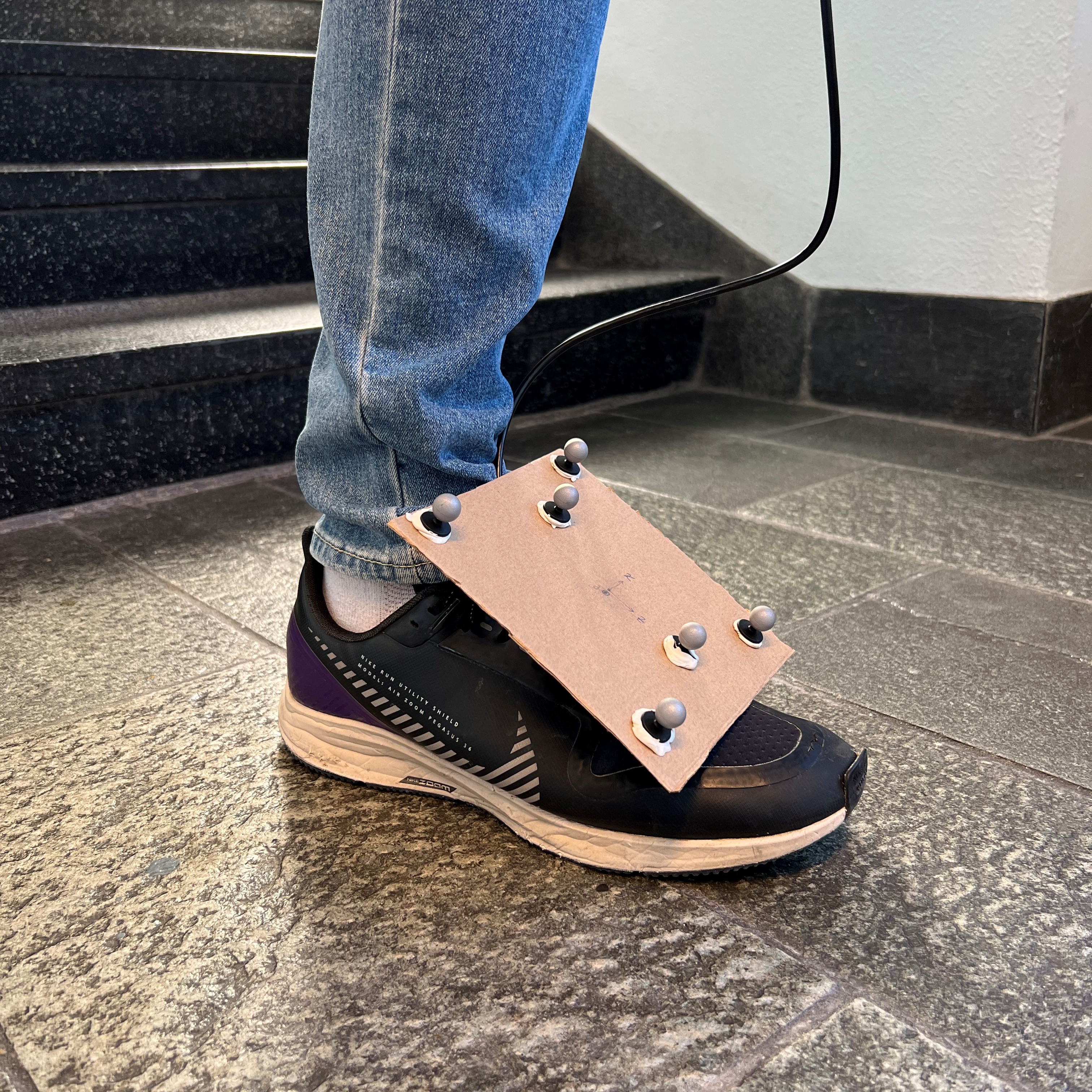} \\
	\end{tabular}
	\caption{The experimental setup used to validate the proposed algorithm (left) and the visual markers fixed on top of the IMU for collecting the groundtruth data (right).}
	\label{fig:setup}
\end{figure}

The experiments were executed using a foot-mounted IMU (MTi-100 Xsens). The ground-truth data were collected using markers fixed on a cardboard sheet fixed on top of the sensor. The experimental setup is shown in Fig.~\ref{fig:setup}.
The foot-mounted odometry was computed using the OpenShoe open source implementation~\citep{nilsson2012foot, nilsson2014foot}. The estimation results when only using the magnetic map or the motion map are also reported. 

For each of the sequences with the ground-truth data, the GP hyperparameters of the magnetic map shown in~\eqref{equ:linear_kernel} and~\eqref{equ:se_kernel} are determined using hyperparameter optimization with the ground-truth poses~\citep{solin2020hilbert}. For the longer sequence, the length scale was set to $\ell = 0.3$, and the magnitude scales were set to $\sigma_{\text{SE}} = 1.0$ and $\sigma_{\text{lin.}} = 0.5$. 

The dimensions of the hexagons were chosen as $r^{\text{m}} = 5 \ \text{m}$, $r^{\text{u}} = 0.5 \ \text{m}$, $L_{z}^{\text{m}} = 2 \ \text{m}$ and $L_{z}^{\text{u}} = 0.125 \ \text{m}$. The number of particles used is $N_\text{p} = 100$, the measurement noise of the magnetometer was set to $\mathcal{R}^{\text{m}} = 0.1 \ \mathbf{I}_{3 \times 3}$ which was also determined using the hyperparameter optimization from the short sequences, the vertical motion probability $p_{v}$ was set to $0.001$, and the process noise covariance matrix used is $\mathcal{Q} = \diag([0.001, 0.001, 0.01, 2\times 10^{-6}, 2\times 10^{-6}, 2\times 10^{-6}])$. The choice of the process noise covariance was based on the output of the ZUPT-EKF odometry which showed more drift in the vertical direction compared to the drift in the horizontal direction. The orientation covariance was chosen to reflect the relatively accurate orientation estimation provided by the ZUPT-EKF algorithm. 

\subsection{Experimental sequences with ground-truth data}
\begin{figure}[t]
	\centering
	\includegraphics[trim={1.0cm 1cm 1.5cm 0cm}, clip, width=0.5\linewidth]{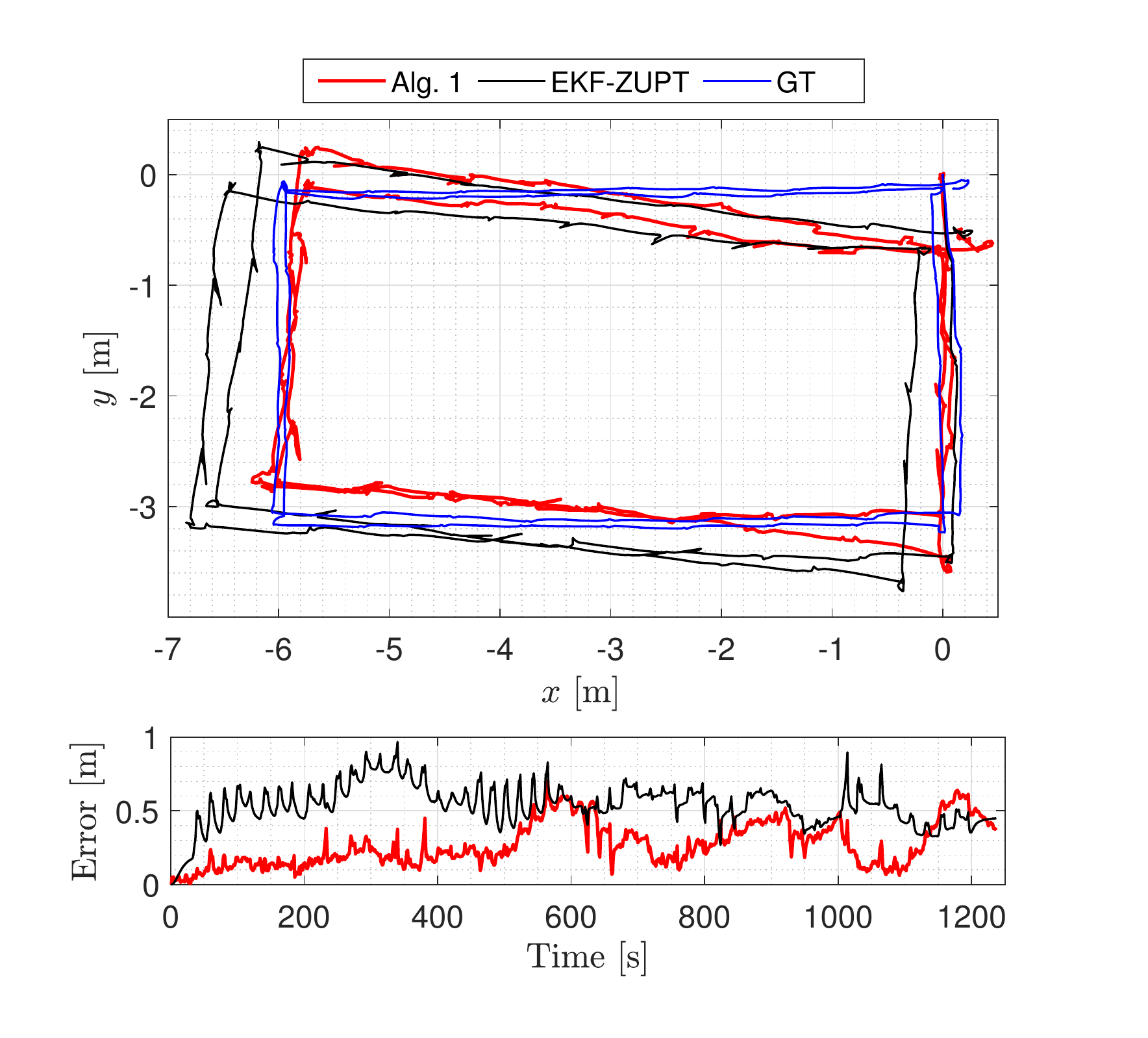}
	\caption{Top: The estimated path of sequence I using the ZUPT-EKF algorithm (black) and Alg.~\ref{alg:LSMHE} (red) plotted along with the ground truth (blue). Bottom: The translation error for the estimated paths by Alg.~\ref{alg:LSMHE} and the ZUPT-EKF.}
	\label{fig:seq1}
\end{figure}

Fig.~\ref{fig:seq1} shows the position estimates as well as the position estimation errors of Alg.~\ref{alg:LSMHE} for sequence I. 
In Fig.~\ref{fig:seq1}, it can be seen that Alg.~\ref{alg:LSMHE} decreases the position estimation error that was present in the odometry. It can also be seen, however, that the error increases slightly in the last 10 seconds, which is due to some inaccuracies in the magnetic field map. 
Alg.~\ref{alg:LSMHE} was still able to reduce the error in the estimated trajectory from a translation RMSE of $0.56$ m in the case of the ZUPT-EKF to $0.41$ m (see Table~\ref{tab:seq1trans}. 

Table~\ref{tab:seq1trans} and~\ref{tab:seq1ori} contain the translation and orientation RMSEs in sequence I for Alg.~\ref{alg:LSMHE} as well as the RMSE for the SLAM algorithm using the magnetic maps only and the motion maps only. For each of the SLAM algorithms, the average and standard deviation of the translation and orientation errors are reported. The proposed SLAM algorithm yielded more accurate pose estimation results both for translation and orientation which shows the complementary performance of both maps in achieving accurate pose estimation results. 

\begin{figure}[t]
	\centering
	\includegraphics[trim={1.5cm 1.5cm 1.5cm 0cm}, clip, width=0.6\linewidth]{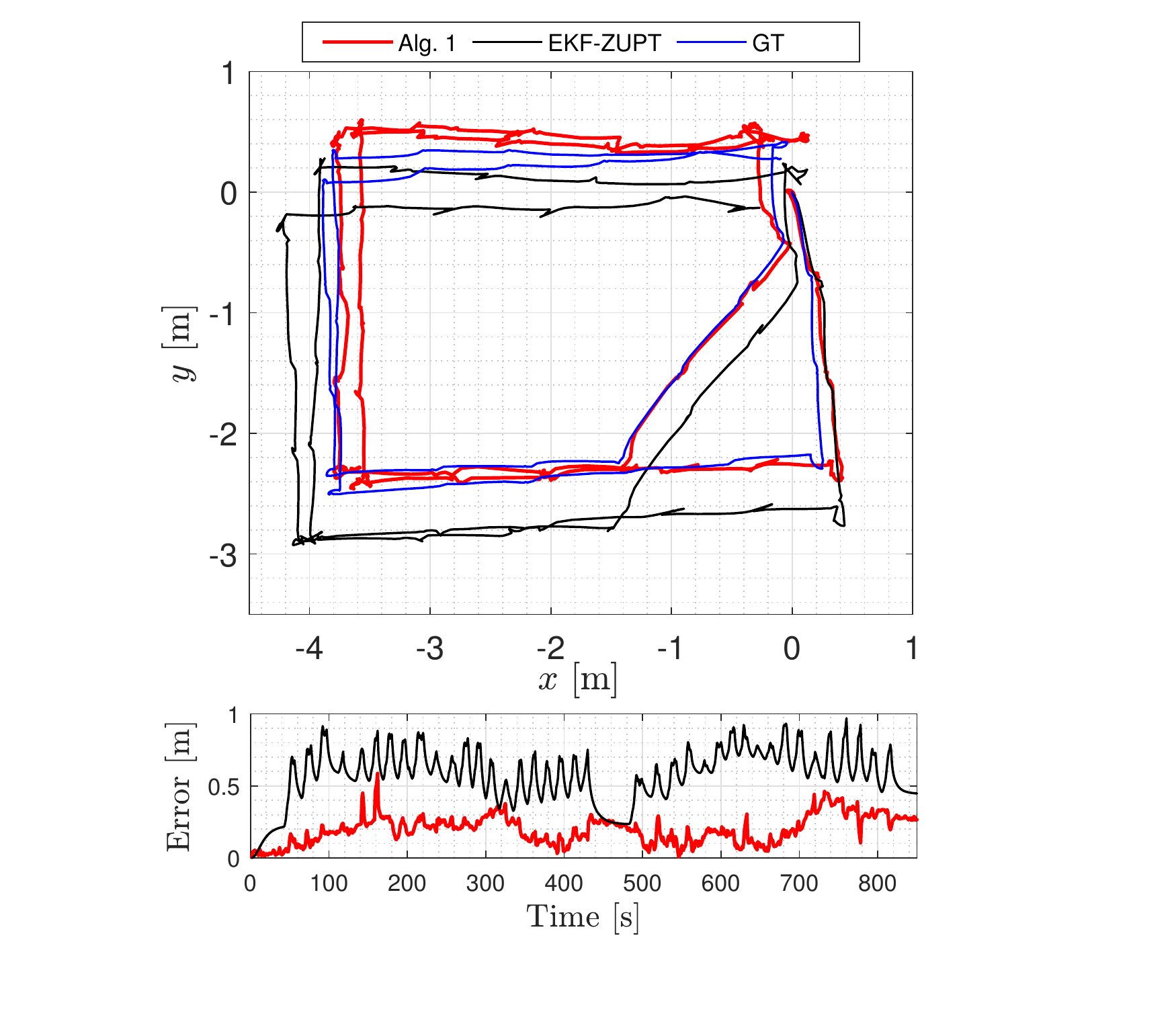}
	\caption{Top: The estimated path of sequence II using the ZUPT-EKF algorithm (black) and Alg.~\ref{alg:LSMHE} (red) plotted along with the ground truth (blue). Bottom: The translation error for the estimated paths by Alg.~\ref{alg:LSMHE} and the ZUPT-EKF.}
	\label{fig:seq2}
\end{figure}

Fig.~\ref{fig:seq2} shows the position estimates as well as the position estimation errors of Alg.~\ref{alg:LSMHE} for sequence II. 
This sequence is composed of two loops (compared to sequence II) with a sloped turn during the second one to test the performance of the algorithm against non-repetitive paths as this makes it more difficult for the motion map to model the motion of the pedestrian. 
Even with the sloped turn, the algorithm was capable of accurately estimating the path of the pedestrian and correctly closing the loop when the pedestrian returned to the original rectangular path taken during the first loop. In sequence II, the translation RMSE was reduced from $0.6$ m in the ZUPT-EKF to $0.32$ m in the case of Alg.~\ref{alg:LSMHE}. 

\begin{table}[ht]
	\centering
	\caption{Translation RMSE For 10 Runs of Sequence I.}
	\label{tab:seq1trans}
	\renewcommand{\arraystretch}{1.2}
	\begin{tabular}{cccc}
		\hline \hline
		RMSE   & Horizontal [m] & Vertical [m] & Total [m] \\ \hline
		Alg. 1 (no mag. map)  & 0.66 $\pm$ 0.19  & 0.09 $\pm$ 0.01    & 0.66 $\pm$ 0.19 \\ \hline
		Alg. 1 (no motion map)  & 0.43 $\pm$ 0.10  & 1.50 $\pm$  0.38    & 1.57 $\pm$ 0.33  \\ \hline
		Alg. 1 & \textbf{0.40} $\pm$ 0.06  & \textbf{0.08} $\pm$ 0.01     & \textbf{0.41} $\pm$ 0.06 \\ \hline \hline
	\end{tabular}
\end{table}

\begin{table}[ht]
	\centering
	\caption{Orientation RMSE For 10 Runs of Sequence I.}
	\label{tab:seq1ori}
	\renewcommand{\arraystretch}{1.2}
	\begin{tabular}{cccc}
		\hline \hline
		RMSE   &  Roll [rad] & Pitch [rad] & Yaw [rad]   \\ \hline
		Alg. 1 (no mag. map)  & 0.137 $\pm$ 0.003 & 0.060 $\pm$ 0.010 & \textbf{0.132} $\pm$ 0.008 \\ \hline
		Alg. 1 (no motion map)  & \textbf{0.136} $\pm$ 0.001 & 0.059 $\pm$ 0.007  & 0.140 $\pm$ 0.018 \\ \hline
		Alg. 1 & \textbf{0.136} $\pm$ 0.001 & \textbf{0.057} $\pm$ 0.004 & \textbf{0.132} $\pm$ 0.006 \\ \hline \hline
	\end{tabular}
\end{table}

Table~\ref{tab:seq2trans} and~\ref{tab:seq2oir} shows the translation and orientation RMSEs for sequence II. 
The motion map is based on the assumption that it's more likely to traverse the same path multiple times. Violation of this assumption, therefore, degrades the results of Alg. 1 without the magnetic map. It can also be seen to have a slightly negative effect on the results from Alg. 1 as compared to not only using the magnetic field map. However, this negative effect can be seen to be small.

Using Alg.~\ref{alg:LSMHE} without the motion map, the position estimation results can be seen to suffer from poor odometry in the vertical direction. On the other hand, using Alg.~\ref{alg:LSMHE} without the magnetic field map can be seen to result in low position accuracy in the horizontal direction. This can be explained by the fact that the accuracy of this algorithm is limited by the size of the hexagonal tiles. Alg.~\ref{alg:LSMHE} can be seen to overcome both limitations and result in a lower RMSE. 

\subsection{Largescale experiment}
Fig.~\ref{fig:PMSLAM} shows a longer testing sequence collected without ground-truth data. The figure shows the overall operation of the proposed SLAM algorithm where the large black hexagons are the hexagonal tiles of the magnetic field map while the smaller blue tiles are the motion map tiles. 
The magnetic field is shown by the heat map. 
Like the underlying assumption in~\citep{kok2018scalable}, there are indeed magnetic field variations present in the environment which are used by Alg.~\ref{alg:LSMHE} as position information.
The sequence was designed so that the starting point (shown as a circular green mark in Fig.~\ref{fig:PMSLAM}) is the same as the end point of the path. As can be seen in the figure, the output of Alg.~\ref{alg:LSMHE} estimated a path that ends closer to the actual end point with a deviation of $0.5$ m compared to the output of the odometry which was $2.1$ m. 

\section{Conclusion and Future Works}
\label{sec:conclusion}
In this paper, an indoor SLAM algorithm was developed for pedestrians using a foot-mounted IMU and magnetic measurements. The algorithm estimates the pose of the pedestrian's foot while building two maps, namely, a magnetic field map and a motion map. The magnetic map models the magnetic field in the environment using Gaussian process regression and magnetometer measurements. The motion map models the map of the indoor environment based on the typical directions of motion of the pedestrian. The algorithm is validated using experimental sequences with the ground-truth data collected using an optical motion capture system as well as a long sequence without the ground-truth data. A comparative study of the performance of the algorithm is performed and the localization results of using only the motion map, the magnetic map, or the proposed algorithm are reported. The results show that the algorithm reduces the drift from the odometry and provides more accurate position and orientation estimates of the pedestrian.

In future work, the motion map can be further extended to model the orientation in addition to the position. Furthermore, the proposed algorithm can be extended to multiple floors scenarios where the motion map would also detect the presence of stairs or elevators in the environment and utilize such information for position and orientation estimation. 

\begin{table}[ht]
	\centering
	\caption{Translation RMSE of 10 Runs of Sequence II.}
	\label{tab:seq2trans}
	\renewcommand{\arraystretch}{1.2}
	\begin{tabular}{cccc}
		\hline \hline
		RMSE   & Horizontal [m] & Vertical [m] & Total [m] \\ \hline
		Alg. 1 (no mag. map)  & 0.52 $\pm$ 0.20       & 0.08   $\pm$ 0.00  & 0.53 $\pm$ 0.20 \\ \hline
		Alg. 1 (no motion map)  & 0.33  $\pm$  0.05    & 1.52 $\pm$ 0.09    & 1.56 $\pm$ 0.08  \\ \hline
		Alg. 1 & \textbf{0.31}   $\pm$  0.06   & \textbf{0.08}  $\pm$  0.01   & \textbf{0.32} $\pm$ 0.06 \\ \hline \hline
	\end{tabular}
\end{table}

\begin{table}[H]
	\centering
	\caption{Orientation RMSE of 10 Runs of Sequence II.}
	\label{tab:seq2oir}
	\renewcommand{\arraystretch}{1.2}
	\begin{tabular}{cccc}
		\hline \hline
		RMSE   & Roll [rad]  & Pitch [rad] & Yaw [rad]   \\ \hline
		Alg. 1 (no mag. map)  & 0.036 $\pm$ 0.008 & 0.032 $\pm$ 0.009 & 0.050 $\pm$ 0.026 \\ \hline
		Alg. 1 (no motion map)  & \textbf{0.032} $\pm$ 0.010 & \textbf{0.028} $\pm$ 0.012 & 0.052 $\pm$ 0.031 \\ \hline
		Alg. 1 & 0.034 $\pm$ 0.005 & 0.031 $\pm$ 0.007 & \textbf{0.046}  $\pm$ 0.017 \\ \hline \hline
	\end{tabular}
\end{table}



\section*{Acknowledgment}
This publication is part of the project ``Sensor Fusion For Indoor Localisation Using The Magnetic Field" with project number 18213 of the research program Veni which is (partly) financed by the Dutch Research Council (NWO). 


\bibliographystyle{IEEEtran}
\bibliography{journal}

\clearpage



\end{document}